\documentclass[letterpaper,11pt]{article} 
\usepackage{fullpage}
\usepackage{mathpazo}
\usepackage[numbers, compress, sort]{natbib}
\usepackage[utf8]{inputenc} 
\usepackage[T1]{fontenc}    
\usepackage{hyperref}       
\usepackage{url}            
\usepackage{booktabs}       
\usepackage{amsfonts}       
\usepackage{nicefrac}       
\usepackage{microtype}      
\usepackage{caption}
\usepackage{amssymb}
\usepackage{graphicx}
\usepackage{subfigure}
\usepackage{threeparttable}
\usepackage{color}
\usepackage{complexity}
\usepackage{amsmath}
\usepackage{amsthm}
\usepackage{listings}
\usepackage{pifont}
\usepackage{tabularx}
\usepackage{fancyvrb}
\usepackage{comment}
\usepackage{algorithm}
\usepackage{algorithmic}
\usepackage{float}
\usepackage{wrapfig}
\usepackage{mathtools}
\usepackage{thmtools,thm-restate}
\usepackage{paralist}
\usepackage{multirow}
\usepackage{scrextend}
\usepackage{enumitem}
\usepackage[noorphans]{quoting}
\usepackage{bm}
\usepackage{bbm}
\usepackage{thm-restate}
\usepackage{tabularx}
\usepackage{xcolor}
\usepackage{xspace}
\usepackage{times}
\usepackage{bbding}

\title{DPNAS: Neural Architecture Search for Deep Learning \\ with Differential Privacy}
\date{}

\author{
Anda Cheng\textsuperscript{\rm 1}, 
Jiaxing Wang\textsuperscript{\rm 2}, 
Xi Sheryl Zhang\textsuperscript{\rm 1}, 
Qiang Chen\textsuperscript{\rm 1}, 
Peisong Wang\textsuperscript{\rm 1}, 
Jian Cheng\textsuperscript{\rm 1} \\
\textsuperscript{\rm 1}Institute of Automation, Chinese Academy of Sciences, \textsuperscript{\rm 2}JD.com \\
\texttt{chenganda2017@ia.ac.cn},
\texttt{\{wjiaxing94, sheryl.zhangxi\}@gmail.com},\\
\texttt{\{qiang.chen, peisong.wang,jcheng\}@nlpr.ia.ac.cn } 
}

\begin{document}

\maketitle

\begin{abstract}
	Training deep neural networks (DNNs) for meaningful differential privacy (DP) guarantees severely degrades model utility. In this paper, we demonstrate that the architecture of DNNs has a significant impact on model utility in the context of private deep learning, whereas its effect is largely unexplored in previous studies. In light of this missing, we propose the very first framework that employs neural architecture search to automatic model design for private deep learning, dubbed as DPNAS. To integrate private learning with architecture search, we delicately design a novel search space and propose a DP-aware method for training candidate models. We empirically certify the effectiveness of the proposed framework. The searched model DPNASNet achieves state-of-the-art privacy/utility trade-offs, e.g., for the privacy budget of $(\epsilon, \delta)=(3, 1\times10^{-5})$, our model obtains test accuracy of $98.57\%$ on MNIST, $88.09\%$ on FashionMNIST, and $68.33\%$ on CIFAR-10. Furthermore, by studying the generated architectures, we provide several intriguing findings of designing private-learning-friendly DNNs, which can shed new light on model design for deep learning with differential privacy.
\end{abstract}

\section{Introduction}\label{sec-intro}

Deep neural networks (DNNs) have achieved massive successes in a variety of tasks, including image understanding, natural language processing, etc  \cite{dlbook, dlnature}.
Broadly, on the other hand, DNNs may compromise sensitive information carried in the training data \cite{mia, inversion, dlg},
thereby raising privacy issues. 
To counter this, learning algorithms that provide principled privacy guarantees in the line of differential privacy (DP) \cite{dp, ethical, dpcomplexity} have been developed.
For instance, differentially private stochastic gradient descent (DPSGD) \cite{dpsgd}, a generally applicable modification of SGD, is widely adopted in differentially private deep learning (DPDL) applications, ranging from medical image recognition \cite{p3sgd}, image generation \cite{dpgan}, to federated learning \cite{fed}, to name a few.

However, training DNNs with strong DP guarantees inevitably degrades model \textit{utility} \cite{dp}. 
To improve the utility for meaningful DP guarantees, prior works have proposed a wealth of strategies, e.g., gradient dimension reduction \cite{gep, bgep}, adaptive clipping \cite{adaclip1, adaclip2}, adaptive budget allocation \cite{adpt-budget}, transfer learning with non-sensitive data \cite{scatter, ppsgd}, and so forth. 
Whilst all these methods focus on improving training algorithms, the impact of model architectures on the utility of DPDL is so far unexplored. 
Formally, a learning algorithm $\mathcal{M}$ that trains models from the output set $\mathcal{O}=\text{Range}(\mathcal{M})$ is $(\epsilon, \delta)$-DP, if $\text{Pr}[\mathcal{M}(D) \in \mathcal{O}] \le e^{\epsilon} \text{Pr}[\mathcal{M}(D') \in \mathcal{O}] + \delta$ holds for all training sets $D$ and $D'$ that differ by exactly one record.
For DPSGD, its output at each step is a high-dimension gradient vector, which is affected by model architecture.
As a result, using different model architectures do not change the privacy of DPSGD but could have effect on the output space $\mathcal{O}$, thereby affecting the model utility.


\begin{figure}[tp]
	\begin{center}
		\includegraphics[width=.650 \linewidth]{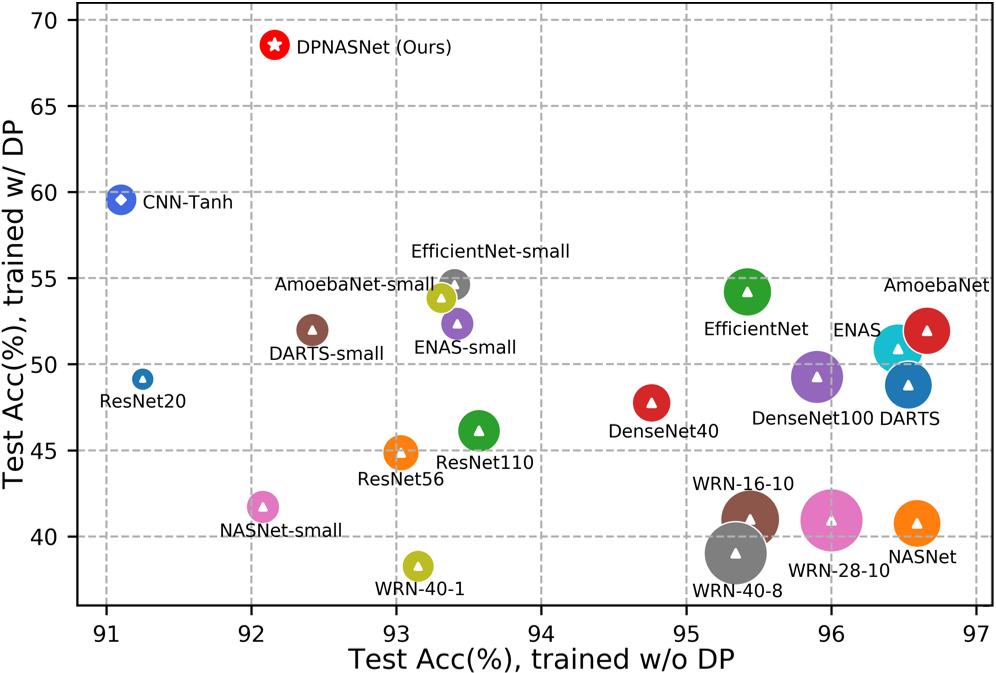}
	\end{center}
	\caption{Comparing the utility of different models for DPDL on CIFAR-10. The y-axis and x-axis show the test accuracy of models trained with DPSGD and SGD, respectively.
	The triangle core denotes the model designed for conventional deep learning. 
	The diamond core denotes the model designed for DPDL.
	The size of the circumcircle reflects the number of model parameters.
	Our DPNASNet is superior to the existing DNN models for DPDL. 
    }
	\label{compare}
\end{figure}

To investigate the effect of model architectures, we delve into the utility comparisons of several existing DNNs by training them with SGD and DPSGD. 
The comparisons include hand-crafted models \cite{res, dense, wrn, ts} and automatically searched models \cite{nasnet, darts, efnet, real19, enas}.
Without loss of generality, three observations can be summarized from Figure \ref{compare}.
Firstly, for DPDL, the models designed without considering private training, e.g. triangle cores, perform much poorer than the model specially designed, e.g. diamond core.
Secondly, there is no clear positive correlation between the performance of a model trained with and without DP.
Lastly, the model utility obtained with DPSGD varies greatly even though the model size are comparable, which appears to be a different trend comparing to the model performance in conventional deep learning.
At a high level, the above observations suggest that (1) \textit{Model architectures can significantly affect the utility of DPDL;} (2) \textit{Instead of directly using DNNs built for conventional deep learning, it is necessary to redesign models for DPDL to improve the utility.}
Whilst model architecture is important for DPDL, there is little experience or common knowledge to draw on to design DP-friendly models.

In this paper, inspired by the above insights and the advances of neural architecture search (NAS) \cite{nas}, we propose DPNAS, the first effort to automatically search models for DPDL.
Our motivation is to boost the privacy/utility trade-offs with least prior knowledge
by integrating private learning with architecture search.
To this end, we design a novel search space for DPDL and propose a DP-aware method for training candidate models during the search process.   
The searched model DPNASNet achieves new SOTA results. 
Especially, for privacy budget $(\epsilon, \delta)=(3, 1\times10^{-5})$, we gain the test accuracy of $98.57\%$, $88.09\%$, and $68.33\%$ on MNIST, FashionMNIST, and CIFAR-10, respectively. 
In ablation studies, we verify the effectiveness of our approach and the merits of the resulted models.
The advantages of DPNAS enables us to not only automatically design better models with little prior knowledge but also summarise some rules about model design for DPDL.
We conduct analysis on the resulted models and provide several new findings for designing DP-friendly DNNs, concluded as (1) SELU \cite{selu} is more suitable for DPDL than Tanh (tempered sigmoid); (2) The activation functions that can retain the negative values could be more effective for DPDL; (3) Max pooling is better than average pooling for DPDL.

Our main contributions are summarized as below:
\begin{itemize}
	\item We propose DPNAS, the first framework that employs NAS to search models for private deep learning.
	We introduce a novel search space and propose a DP-aware method to train the candidate models during the search.
	
	\item  Our searched model DPNASNet substantially advances SOTA privacy/utility trade-offs for private deep learning.

	\item We conduct analysis on the resulted model architectures and provide several new findings of model design for deep learning with differential privacy.
\end{itemize}

\section{Related Work}\label{sec-related}

\paragraph{Differentially Private Deep Learning.}
Differential privacy (DP) \cite{dp} is a measurable definition of privacy that provides provable guarantees against individual identification in a data set. 
To address the privacy leakage issue in deep learning, Abadi et al.\cite{dpsgd} propose  differentially private stochastic gradient descent (DPSGD), which is a generally applicable modification of SGD.
Despite providing a DP guarantee, DPSGD brings about a significant cost of model utility.
To ensure the utility of DPSGD trained models, the line of works \cite{gep,bgep,adaclip1,adaclip2,adpt-budget,ppsgd} make efforts to improve the DP training algorithm while neglecting the benefit of enlarging the algorithm space by varying model architectures.
Recently, Papernot et al. \cite{ts} observes that DPSGD can cause exploding model activations, thereby leading to the degradation of model utility.
To make models more suitable for DPSGD training, 
they replace the unbounded ReLUs with the bounded Tanhs as the activation function.
However, their design is hand-crafted and does not involve other factors of model architectures that may also affect the utility.
On the contrary, our work explicitly and systematically studies the effect of model architecture on the utility of DPDL and propose to boost the utility via automatically searching by considering both activation function selection and network topology.

\paragraph{Neural Architecture Search (NAS).}
NAS is an automatic model architecture designing process to facilitate fewer human efforts and higher model utilities. 
A typical process of NAS can be described as follows.
The \textit{search strategy} first samples a candidate model from the \textit{search space}, where 
the model is trained and evaluated according to the \textit{performance estimation}. 
Then, the estimation result is feedback to the search strategy as guidance to select better models. 
Hence, the optimal model could be obtained by sequential iterations.
Following this paradigm, widely used search strategies including reinforcement learning (RL) \cite{nas, nasnet, metaqnn, enas}, evolution algorithms (EA) \cite{real17, real19}, and gradient-based optimization of architectures \cite{darts, mdenas, pnas} are developed. 
For providing a subtle search space, previous methods tend to search optimal connections and operations in the cell level \cite{nasnet, darts, enas, real19, pnas, mdenas}.
By doing so, the overall architecture of a model is constructed with cells that share the same architecture.
The search process attempts to find optimal internal connections and operations of the cell.
The connection topology of the cell can be established based on predefined motifs \cite{hienas} or automatically searched structure \cite{nasnet}.
The operations normally are set as non-parametric operations and convolutions with different filter sizes. 
It deserves to note that the choice of activation functions is usually not taken into account in search spaces dealing with non-privacy-preserving deep learning tasks.
In contrast to the existing NAS methods, our DPNAS aims at searching high-performance models for private learning with a different training method for candidate models on a sophisticated designed search space involving multiple types of activation functions. 


\section{Methodology}\label{sec-method}

Our goal is to employ NAS to search networks that are more suitable for private training.
As the previous NAS methods are proposed for non-privacy-preserving tasks, their formulations are not ideal for achieving our goal.
We introduce our re-formulation in Section \ref{formulation}.
Based on our formulation, the design of our DPNAS framework includes three parts, which are the design of search space, the DP-aware training method for candidate networks, and the search algorithm. 
For the remainder of this section, we present our novel search space in Section \ref{sec-space}, describe the proposed DP-aware training method in Section \ref{train_alg}, and introduce the search algorithm in Section \ref{sec-search}.  

\begin{figure}[tbp]
	\begin{center}
		\includegraphics[width=0.75 \linewidth]{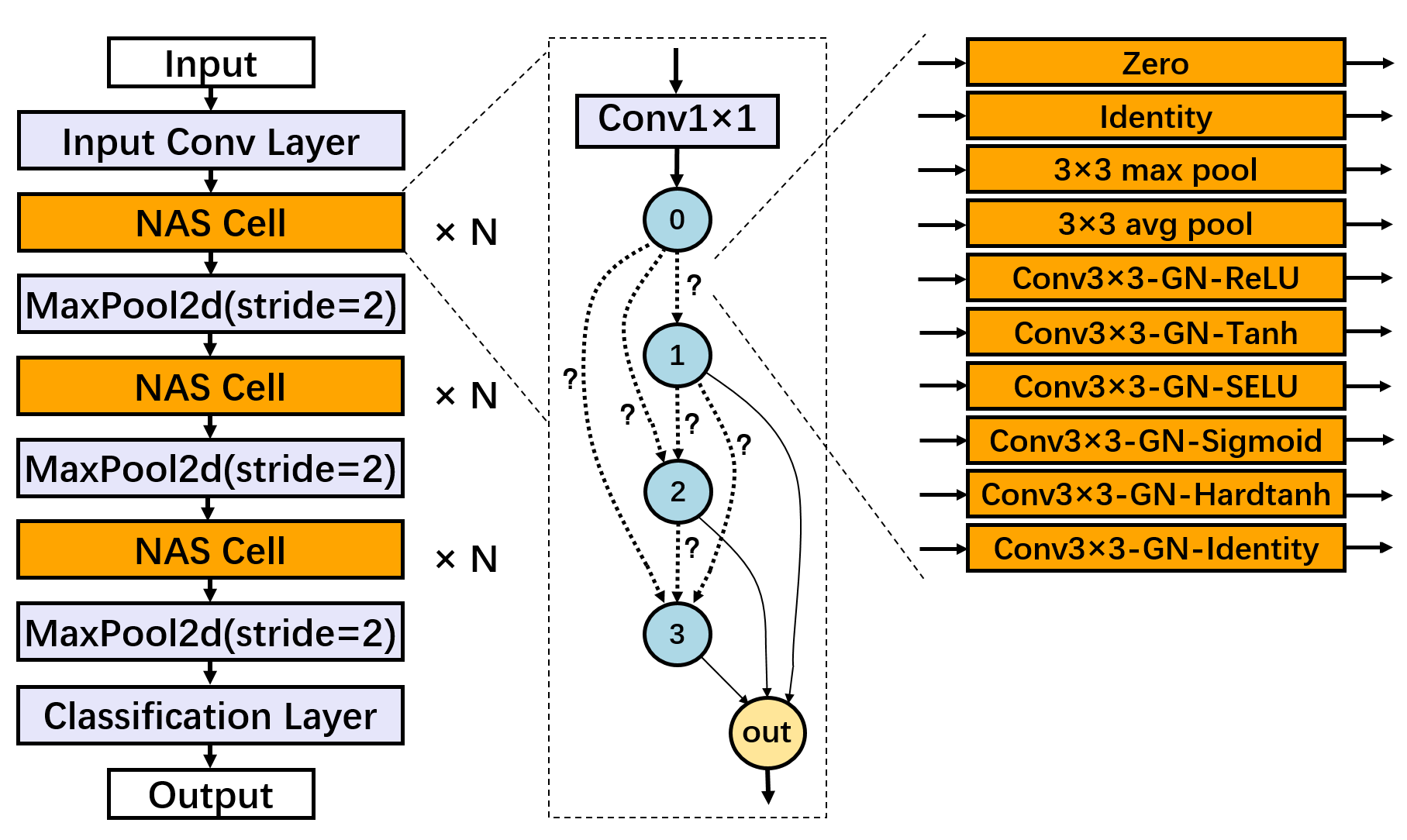}
	\end{center}
	\caption{Illustration of search space. 
	The leftmost part shows the overall model architecture. 
	The middle part shows the predefined fully-connected structure of the NAS cell.
	$"?"$ indicates the operation on each path will be chosen by searching. 
	The rightmost part shows the candidate operations.
	}
	\label{arch}
\end{figure}

\subsection{DPNAS Formulation}\label{formulation}

NAS is formulated as follows. The architecture search space $\mathcal{A}$ is a set of candidate architectures that can be denoted by directed acyclic graphs (DAG). 
For a specific architecture $a \in \mathcal{A}$, the corresponding network can be denoted by $\mathcal{F}(a, w(a))$, where $w$ represents the network weights of the candidate architecture $a$.
The search of the optimal architecture $a^{*}$ in NAS can be formulated as a bi-level optimization:
\begin{align}\label{eq1}
& a^{*}=\max _{a \in \mathcal{A}} \mathcal{R}(\mathcal{F}(a, \mathbf{w}^{*}_a), D_{val}) 
\notag \\ 
\text{s.t. }  & \mathbf{w}^{*}_a=\mathcal{OPT}(\mathcal{F}(a, \mathbf{w}_a), D_{train})
\end{align}
where $\mathcal{R}(\mathcal{F}(a, \mathbf{w}^{*}_a), D_{val})$ is the reward of network $\mathcal{F}(a, \mathbf{w}^{*}_a)$ on validation data $D_{val}$, $\mathcal{OPT}$ is an optimization algorithm to find optimal weight $\mathbf{w}^{*}$ on training data $D_{train}$ given candidate architecture $a$. 
For traditional architecture searching without considering private training,
$\mathcal{OPT}$ is typically selected as a non-private learning algorithm, e.g. SGD.
But for private learning, the resulting architecture $a^{*}$ will be trained by a differentially private learning algorithm such as DPSGD. 
As a result, the architectures resulted by the above formulation is not suitable.
Instead, while considering private-training process, we should replace $\mathcal{OPT}$ in equation \ref{eq1} with a differentially private optimization algorithm $\mathcal{DP\raisebox{0mm}{-}OPT}$, which results to DPNAS formulation as following:
\begin{align}\label{eq2}
& a^{*}=\max _{a \in \mathcal{A}} \mathcal{R}(\mathcal{F}(a, \mathbf{w}^{*}_a), D_{val}) \notag \\
\text{s.t. }  &\mathbf{w}^{*}_a=\mathcal{DP\raisebox{0mm}{-}OPT}(\mathcal{F}(a, \mathbf{w}_a), D_{train})
\end{align}

\subsection{Search Space Design}\label{sec-space}
We specially design a novel search space for DPDL.
Following \cite{nasnet, darts, enas}, we search for computation cell as the basic building unit to construct the whole network architecture.
As shown in the leftmost part of Figure \ref{arch}, the overall structure of the network is chained.
After an input convolution layer, $N$ cells are stacked as main computation blocks. 
All the cells share the same architecture.
The resolution of input features and output features of a cell is the same.   
After each cell, a 3$\times$3 max pooling layer with stride 2 is stacked as a down-sampling layer.  
The stacked cells with a down-sampling layer are referred to as a \textit{block}.
We repeat stacking \textit{blocks} for three times and end up with a classification layer.

We aim at searching the internal connection of the cell.
As shown in the middle part of Figure \ref{arch}, a cell is a fully-connected directed acyclic graph $G = (V,E)$ consisting of an ordered sequence of $K+1$ nodes. 
Each node $V^{i}$ (denoted as blue circle) corresponds to an intermediate feature map
$f^{i}$.
The input node $V^{0}$ is obtained by applying a convolution with filter size $1\times1$ to the inputs, which aims at making the dimension of inputs adapt to the channel size of convolution filters in this cell.  
The nodes $\{V^{i}| {i \in 0...K} \}$ are internal nodes, each of which is connected to all the previous nodes in this cell.
Each directed edge $E^{(i,j)}$ is associated with some operation $o(i;j)$ chosen from a pre-defined operation pool $\mathcal{O}=\{o_k(\cdot)|{k \in 1...n} \}$ containing $n$ candidate operations.
Each internal node is computed based on all of its predecessors: $f^{(j)}=\sum_{i<j} o^{(i, j)}(f^{(i)})$.
The output of the cell is obtained by applying concatenation to all the internal nodes.

As shown in the rightmost part of Figure \ref{arch}, our operation pool $\mathcal{O}$ is designed with involving different types of activation functions which is much different from the existing NAS methods.
As mentioned in \cite{ts}, the choice of activation functions is crucial for improving the model utility for DPDL.
Therefore, in our search space, we include 5 different types of non-linear activation functions, which are ReLU, Tanh, Sigmoid, Hardtanh, and SELU \cite{selu}, as well as identity as a linear activation function.
The activation functions are integrated into \textit{Conv3$\times$3-Normalization-Activation} blocks.
We use Group Normalization (GN) \cite{gn} as normalization method because the widely used Batch Normalization \cite{bn} could cost privacy budget while GN not \cite{scatter}.
We also take the topology of cells into considering, which is not considered in \cite{ts}.
We include 4 non-parameter operations into our search space, which are \textit{Identity}, \textit{3$\times$3 max pooling}, \textit{3$\times$3 average pooling}, and \textit{Zero}.
The involvement of these operations enables more possible topologies of the cell.
For example, DPNAS can use \textit{Zero} operation to indicate a lack of connection between two nodes in the cell.

\subsection{DP-aware Training }\label{train_alg}

A differently private learning algorithm execute different process compared with its corresponding non-private algorithm.
With considering this difference, we need to choose a private training algorithm as $\mathcal{DP\raisebox{0mm}{-}OPT}$ in equation \ref{eq2} for training the sampled candidate networks to guide the resulted networks more adaptive to $\mathcal{DP\raisebox{0mm}{-}OPT}$.
In our implementation, we use DPSGD as $\mathcal{DP\raisebox{0mm}{-}OPT}$.
DPSGD makes two changes to every iteration of SGD before updating model weights with computed gradients.
It firstly bounds the sensitivity of the learning process to each individual training example by computing per-example gradients $\{g_i| {i \in \{ 0...m-1}\}$ with respect to the training loss for $m$ model parameters $\{w_i| {i \in 0...m-1}\}$, and clipping each per-example gradients to a maximum fixed $l_2$ norm $C$.
DPSGD then adds Gaussian noise $N(0, \sigma^2C^2\text{I})$ to the average of these per-example gradients, where $\sigma$ is noise intensity selected according to the privacy budget $(\epsilon, \delta)$ \cite{dpsgd}. 
Previous works \cite{ts, gep} indicate that both of these two steps could make negative impacts on the learning process, thereby degrading the utility of the resulted models.

To make the search process aware of the impact of those two steps, we include those two steps into the training processes of sampled architectures and carefully select the value of hyper-parameters $C$ and $\sigma$ for the search.
In practice, we tend to set $C$ to be small as the intensity of the added noise scales linearly with $C$.
To make the resulted architectures more adaptive to the gradient clipping, in the search process, we set $C$ small, e.g. $C=0.1$.
As for $\sigma$, we can choose $\sigma$ without considering privacy cost during the search processes because the resulted architectures will be trained on private datasets from scratch.
In practice, we empirically find that setting $\sigma$ to large values could severely slow down the convergence rate of sampled architectures, thereby making the search process unreliable. 
Therefore, in our implementation, we set $\sigma$ to be relatively small.

\renewcommand{\algorithmicrequire}{\textbf{Input:}}
\renewcommand{\algorithmicensure}{\textbf{Output:}}
\begin{algorithm}[t]
    \caption{Search Process of DPNAS}
    \label{search-alg}
	\begin{algorithmic}[1]
		\REQUIRE Training set $D_{train}$, validation set $D_{val}$, batch size $B_n$ for training candidate networks, batch size $B_c$ for training controller, learning rate $\eta_n$ for networks, learning rate $\eta_c$ for controller, network training iterations $T_n$, controller training iterations $T_c$, total epochs $E$.  
		\ENSURE Trained controller $S_\theta$
		\STATE Initialize controller $S_\theta$; 
		\STATE Initialize weights of modules in search space $\mathcal{A}$; 
		\FOR{$e=1$ to $E$}
		\FOR{$i=1$ to $T_n$} 
		\STATE Sample a batch $\{d_i\}_{i=1}^{B_{n}} \subseteq D_{train}$;
		\STATE Sample a model architecture $a \in \mathcal{A}$ using $S_{\theta}$;
		\STATE Train model:
			 $\mathbf{w}_a \! \leftarrow \! \mathcal{DP\raisebox{0mm}{-}OPT}(a,\mathbf{w}_a,\{d_i\}_{i=1}^{B_{n}})$;
		\ENDFOR
		\FOR{$i=1$ to $T_c$}
		\STATE Sample a batch $\{d_i\}_{i=1}^{B_c} \subseteq D_{val}$;
		\STATE Sample an architecture $a \in \mathcal{A}$ using $S_{\theta}$;
		\STATE Update controller:
    		$\theta \leftarrow \theta + \eta_{c} \cdot \frac{1}{B_c} \sum_{i} \nabla_{\theta} \mathcal{R}(a, d_i)$;
		\ENDFOR
		\ENDFOR
		\RETURN $S_\theta$
	\end{algorithmic}
\end{algorithm}

\subsection{Search Strategy}\label{sec-search}
We use RL-based search strategy \cite{nas, nasnet, enas} with parameter sharing \cite{enas}.
Given the number of internal nodes $K$, at each step of the search, the RNN controller samples an architecture denoted as a operation sequence $\{o^{(i,j)}| {i \in 0...K}, {j \in 0...K}, i<j  \}$.
This architecture is trained on a batch of training data using the training method described in Section \ref{train_alg}.
After repeating the above sampling and training step for an epoch on the training set, the RNN controller is then trained with the policy gradient on the validation set aiming at selecting architectures that maximizes the expected reward.
The expected reward is defined as the validation accuracy of a network adding the weighted controller entropy.
The search process executes the above two processes alternately until the RNN controller achieves convergence.
The overall search process is depicted in Alg. \ref{search-alg}.

\section{Experiments}\label{sec-exp}

\subsection{Experimental Settings}
We run DPNAS on MNIST \cite{mnist}, FashionMNIST \cite{fmnist}, and CIFAR-10 \cite{cifar}.
We split the training data of each dataset into the training set and validation set with the ratio of $0.6:0.4$.
For the overall architecture, the number of stacked NAS cells in each stage is $N=1$, the number of internal nodes is $K=5$, and the internal channels of the three stages are 48, 96, 192 for CIFAR-10 and 32, 32, 64 for MNIST and FashionMNIST.
The sampled architectures are trained with DPSGD with weight decay $2\times10^{-4}$, and moment 0.9.
The batch size is set to 300 and the learning rate is set to $0.02$.
The RNN controller used in our search process is the same as the RNN controller used in \cite{enas}.
It is trained with Adam optimizer \cite{adam}. 
The batch size is set to 64 and the learning rate is set to $3\times10^{-4}$.
The trade-off weight for controller entropy in the reward is set to 0.05.
Our search process runs for 100 epochs. 
The first 25 epochs are warm-up epochs that only training sampled architectures without updating the RNN controller.

We evaluate the utility of searched models for DPDL on three common benchmarks: MNIST \cite{mnist}, FashionMNIST \cite{fmnist}, and CIFAR-10 \cite{cifar}. 
On each dataset, the evaluated models are constructed by the resulted cells searched on this dataset.
The evaluated models are trained on the training set \textit{from scratch} using DPSGD and tested on the testing set. 
The privacy cost of training is computed by using the Rényi DP analysis of Gaussian mechanism \cite{rdp, srdp}.
We implement the search process and private training by PyTorch \cite{torch} with \textit{opacus} package.
As for model architectures, the number of stacked NAS cells in each stage is $N=1$, the number of internal nodes is $K=5$, and the internal channels of three stages are set to 48, 96, 192 for CIFAR-10 and 32, 32, 64 for MNIST and FashionMNIST.
All experiments are conducted on a NVIDIA Titan RTX GPU with 24GB of RAM.

\begin{table*}[tbh]
	\centering  
	\small
	\begin{tabular}{@{}llccccccccccccc@{}}
		\toprule
		\multirow{2}{*}{Datasets}
		 & & \multicolumn{3}{c}{Models} \\
		\cmidrule(l){3-5} 
		 & & CNN-Tanh\cite{ts} & DPNASNet-S & DPNASNet \\

		\midrule
		\multirow{4}{*}{MNIST} 
		& Acc($\%$), $\epsilon= 1$ & $89.08\pm0.07$ & $92.87\pm0.07$ & $\mathbf{97.22}\pm0.08$ \\
		& Acc($\%$), $\epsilon= 2$ & $97.24\pm0.10$ & $97.74\pm0.08$ & $\mathbf{98.18}\pm0.11$ \\
		& Acc($\%$), $\epsilon= 3$ & $98.10\pm0.07$ & $98.35\pm0.09$ & $\mathbf{98.57}\pm0.10$ \\
		& $\#$Params & $0.03$M & $0.03$M & $0.21$M \\

		\midrule
		\multirow{4}{*}{FashionMNIST} 
		& Acc($\%$), $\epsilon= 1$ & $73.86\pm0.12$ & $78.66\pm0.14$ & $\mathbf{82.08}\pm0.28$ \\
		& Acc($\%$), $\epsilon= 2$ & $82.03\pm0.17$ & $84.83\pm0.16$ & $\mathbf{86.17}\pm0.28$ \\
		& Acc($\%$), $\epsilon= 3$ & $86.25\pm0.20$ & $86.73\pm0.19$ & $\mathbf{88.09}\pm0.29$ \\
		& $\#$Params & $0.03$M & $0.03$M & $0.21$M   \\

		\midrule
		\multirow{4}{*}{CIFAR-10} 
		& Acc($\%$), $\epsilon= 1$ & $38.74\pm0.27$ & $-$ & $\mathbf{52.95}\pm0.35$ \\
		& Acc($\%$), $\epsilon= 2$ & $52.74\pm0.25$ & $-$ & $\mathbf{65.03}\pm0.42$ \\
		& Acc($\%$), $\epsilon= 3$ & $59.07\pm0.28$ & $-$ & $\mathbf{68.33}\pm0.45$ \\
		& $\#$Params & $0.55$M & $-$  & $0.53$M   \\

		\bottomrule
	\end{tabular}
	\setlength{\abovecaptionskip}{5pt}
	\setlength{\belowcaptionskip}{5pt}
	\caption{Comparison with the SOTA model for DPDL on MNIST, FashionMNIST, and CIFAR-10. DPNASNet-S is a smaller version of DPNASNet, it has comparable parameters with CNN-Tanh. 
	The models are trained with DPSGD for the privacy budgets of $\epsilon=1,2,3$ and $\delta=1\times10^{-5}$.} 
    \label{comp-sota}  
\end{table*}

\begin{table*}[htb]
	\centering  
	\small
	\begin{tabular}{@{}lcccccccccccccc@{}}
		\toprule
		\multirow{2}{*}{Models}
		& \multirow{2}{*}{$\#$Params}
		& \multicolumn{3}{c}{Acc(\%) under budget $\epsilon$} \\
		\cmidrule(l){3-5} 
		& & $\epsilon=1$ & $\epsilon=2$ & $\epsilon=3$ \\
		\midrule
		NASNet \cite{nasnet} 
		& $3.3$M & $26.29\pm0.45$ & $33.44\pm0.45$ & $40.77\pm0.51$    \\
		NASNet-S$^{\dagger}$ \cite{nasnet}
		& $0.6$M & $24.51\pm0.32$ & $36.56\pm0.40$ & $41.72\pm0.48$   \\
		AmoebaNet \cite{real19} 
		& $3.2$M & $39.21\pm0.43$ & $46.36\pm0.47$ & $51.46\pm0.49$ \\
		AmoebaNet-S $^{\dagger}$ \cite{real19} 
		& $0.5$M & $38.47\pm0.32$ & $49.01\pm0.46$ & $53.84\pm0.46$ \\
		DARTS \cite{darts} 
		& $3.3$M & $36.08\pm0.46$ & $43.50\pm0.52$ & $48.79\pm0.50$    \\
		DARTS-S$^{\dagger}$ \cite{darts} 
		& $0.6$M & $37.48\pm0.29$ & $47.91\pm0.42$ & $51.99\pm0.42$    \\
		EfficientNet \cite{efnet} 
		& $3.6$M & $40.17\pm0.42$ & $48.93\pm0.45$ & $54.20\pm0.48$ \\
		EfficientNet-S$^{\dagger}$ \cite{efnet}
		& $0.6$M & $41.99\pm0.32$ & $49.76\pm0.36$ & $54.63\pm0.43$ \\
		\midrule
		DPNASNet 
		& $0.5$M & $\mathbf{52.95}\pm0.35$ & $\mathbf{65.03}\pm0.42$ & $\mathbf{68.33}\pm0.45$  \\
		\bottomrule
	\end{tabular}
	\caption{Comparison with existing NAS searched models on CIFAR-10. $\dagger$ indicates the models that have comparable number of parameters with DPNASNet. The models are trained with DPSGD for the privacy budgets of $\epsilon=1,2,3$ and $\delta=1\times10^{-5}$.} 
	\label{comp-nas}  
\end{table*}

\subsection{Main Results}\label{main-result}

In Table \ref{comp-sota}, we compare the searched model DPNASNet with the recent SOTA model CNN-Tanh from \cite{ts}, which is a CNN model with Tanh as the activation function.
Comparisons are conducted on MNIST, FashionMNIST, and CIFAR-10 for different privacy budgets of $\epsilon=1, 2, 3$ with fixed privacy parameter $\delta=1\times10^{-5}$.
On each dataset, DPNASNet is built by stacking the  searched cell on this dataset.
For fair comparison, we also construct a smaller version of DPNASNet by reducing the channel numbers of convolution layers, denoted as DPNASNet-S, which has a comparable number of parameters with CNN-Tanh.
We train each model with DPSGD from scratch for 10 times and report the mean and standard deviation  of test accuracy of 10 models. 
Our DPNASNet achieves new SOTA.
On MNIST, DPNASNet achieves $98.18\%$ for the privacy budget of $\epsilon=2.0$, whereas the previous SOTA reported in \cite{ts} is $98.1\%$ for the privacy budget of $\epsilon=2.93$. 
On CIFAR-10, we match the best accuracy in \cite{ts}, namely $66.2\%$ for the privacy budget of $\epsilon=7.53$, with a much smaller budget of $2.20$, which is an improvement in the DP-guarantee of $e^{5.33} \approx \mathbf{206}$.
We also consider training DPNASNet for this larger DP budget and we get $\mathbf{72.57\%}$ for the privacy budget of $(\epsilon, \delta)=(7.53, 1\times10^{-5})$.

\begin{figure}[thb]
	\centering
	\subfigure[Effect of search space.]{
		\label{space}
		\includegraphics[scale=0.3]{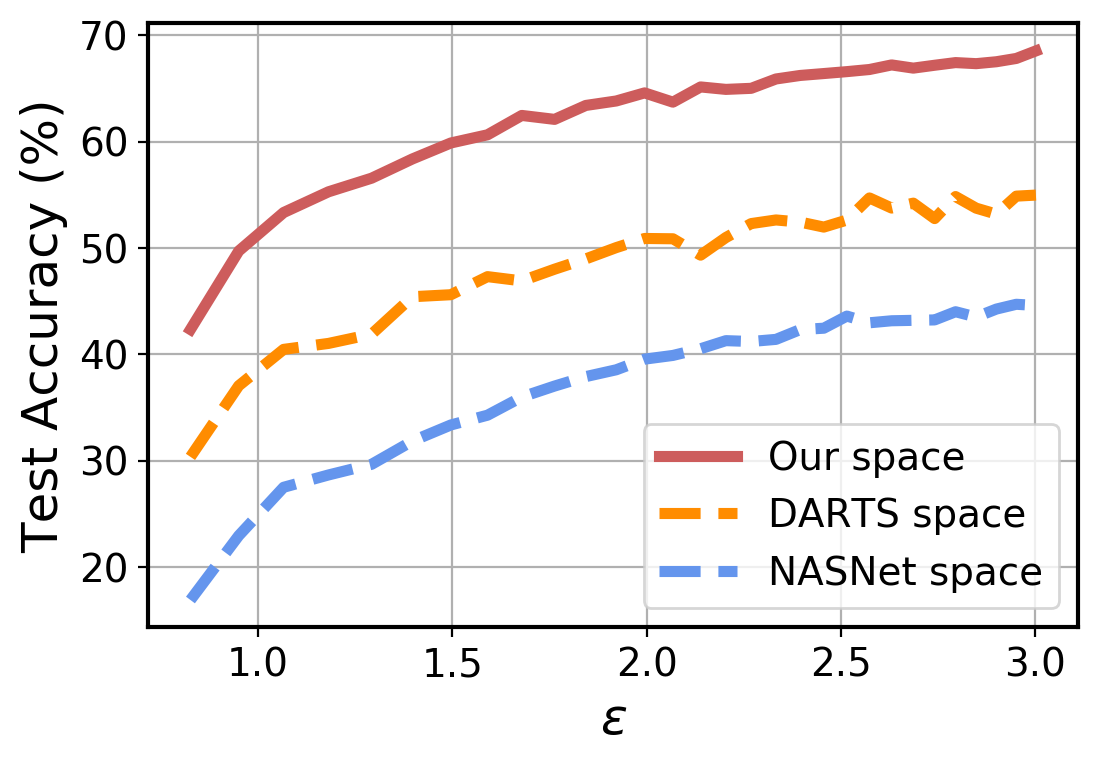}
	}
	\hspace{-0.3cm}
	\subfigure[Effect of training method.]{
	\includegraphics[scale=0.4]{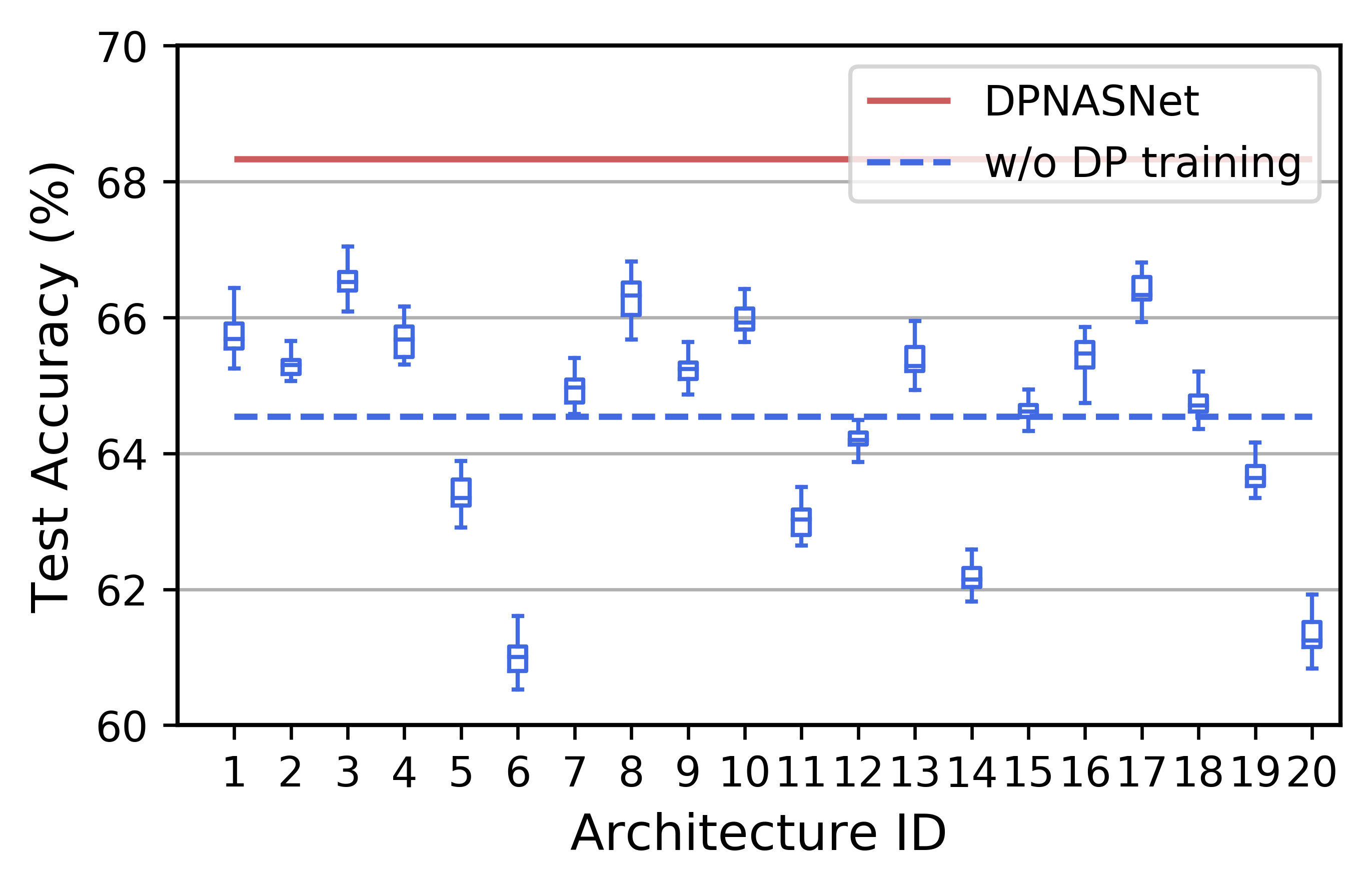}
	\label{train}
	}
	\hspace{-0.3cm}
	\subfigure[Effect of search strategy.]{
		\includegraphics[scale=0.3]{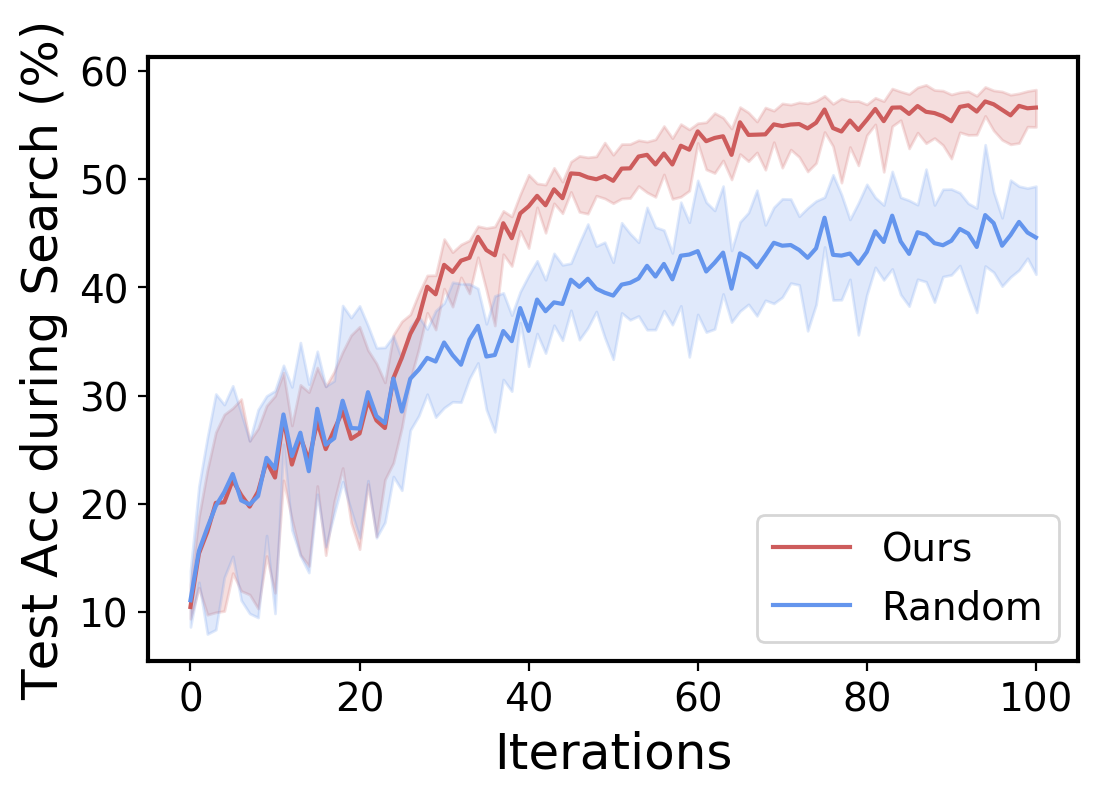}
		\label{search}
	}
	\caption{Ablation studies on CIFAR-10.
	(a) Comparing the utility of models searched on our search space, the search space of DARTS, and the search space of NASNet, using our search process. 
	(b) Comparing the utility between DPNASNet and the models searched without using our training method. The models are trained for the privacy budget of $(\epsilon, \delta)=(3, 1\times10^{-5})$. 
	(c) Comparison of the test accuracy during the search process of RL-based search and random search.}
	\label{ablation}
\end{figure}

\begin{table}[thb]
	\centering  
	\small
	\begin{tabular}{@{}lccc@{}}
		\toprule
		\multirow{2}{*}{Search Dataset} 
		& \multicolumn{3}{c}{Evaluation Dataset (Acc, $\%$)}          
		\\ \cmidrule(l){2-4} 
		& MNIST & FashionMNIST & CIFAR-10 \\ \midrule
		MNIST  
		& $98.57$ & $87.92$ & $67.95$ \\	
		FashionMNIST   
		& $98.58$ & $88.09$ & $68.06$  \\
		CIFAR-10 
		& $98.66$ & $87.94$ & $68.33$ \\
		\bottomrule
	\end{tabular}
	\caption{Transferability of resulted architectures. Each model is constructed by using the cells searched on one dataset (search dataset) and evaluated on other two datasets (evaluation datasets). 
	The models are trained with DPSGD for a privacy budget of $(\epsilon, \delta)=(3, 1\times10^{-5})$.}
	\label{transfer}  
\end{table}

In Table \ref{comp-nas}, we compare DPNASNet with the models searched by the previous NAS methods that do not consider private learning, e.g. DARTS \cite{darts}, NASNet \cite{nasnet}, AmoebaNet \cite{real19}, and EfficientNet \cite{efnet}.
We also compare DPNASNet with the small version of these models  that have a comparable number of parameters with DPNASNet.
The small models are  constructed by reducing the channel numbers of convolution layers in the original models.
We train these models on CIFAR-10 with DPSGD for different privacy budgets of $\epsilon=1,2,3$ with $\delta=1\times10^{-5}$.
From Table \ref{comp-nas}, we observe that DPNASNet dramatically superior to the existing models obtained by NAS methods.

\subsection{Ablation Studies}
We present the ablation study results to verify the effectiveness of our search space, training method, and search strategy. More experimental results are provided in Appendix.

\paragraph{Effect of search space.}
To verify the effectiveness of the search space, we replace our search space with two search spaces widely used in previous NAS works, which are NASNet search space \cite{nasnet} and DARTS search space \cite{darts}.
In NASNet, the predefined cell structure is different from ours.
The cells are divide into two groups, e.g. Normal Cell and Reduction Cell.
In a cell, each internal node has two inputs selected from the outputs of previous nodes or previous two layers. 
In DARTS, the cells are also divided into two groups as NASNet but its predefined cell structure is the same as ours. 
The candidate operations in both of these two search spaces are much different from ours as they do not involve different types of activation functions.
More details about these two search spaces could be found in \cite{nasnet} and \cite{darts}.
We apply our search process on these two spaces and compare the resulted models with DPNASNet.
As shown in Figure \ref{space}, DPNASNet is superior to the models searched on these two spaces, which indicates that our search space is more effective.

\begin{figure*}[tbh]
	\centering
	\subfigure[Searched on MNIST]{
		\includegraphics[scale=0.35]{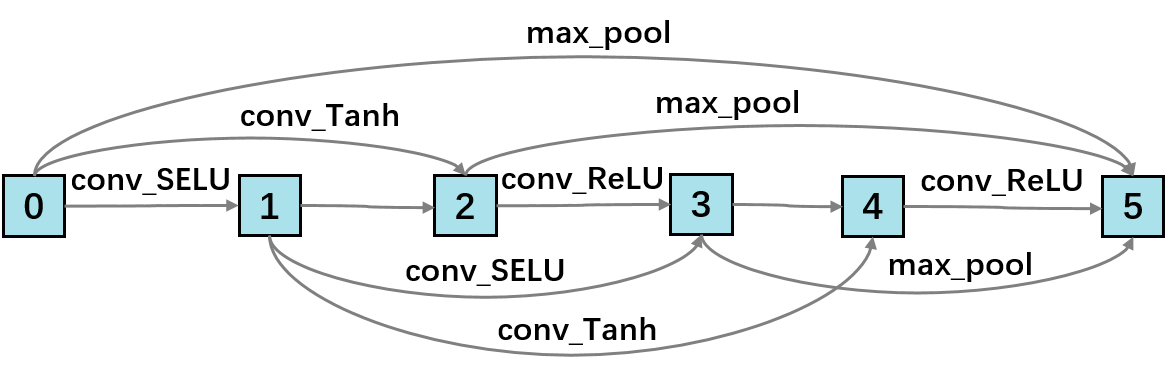}
		\label{cell-mnist}
	}
	\subfigure[Searched on FashionMNIST]{
		\includegraphics[scale=0.35]{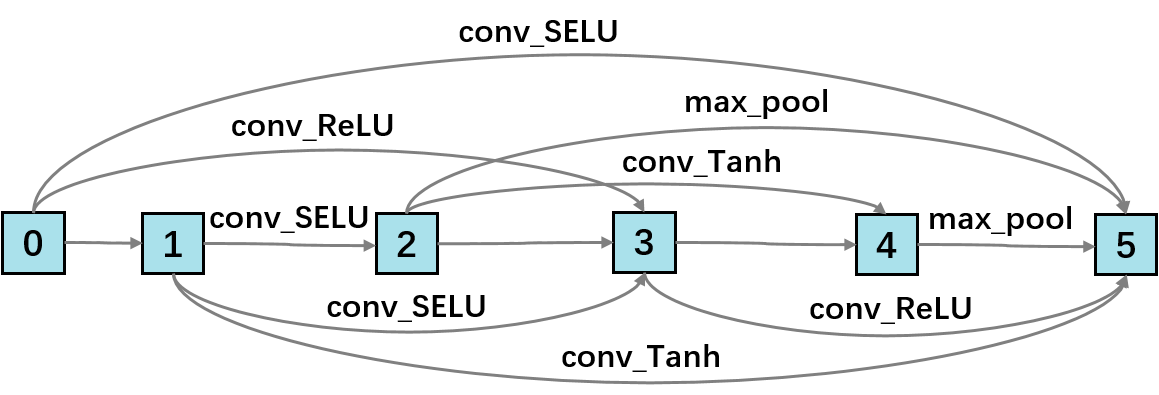}
		\label{cell-fmnist}
	}
	\subfigure[Searched on CIFAR-10]{
		\includegraphics[scale=0.35]{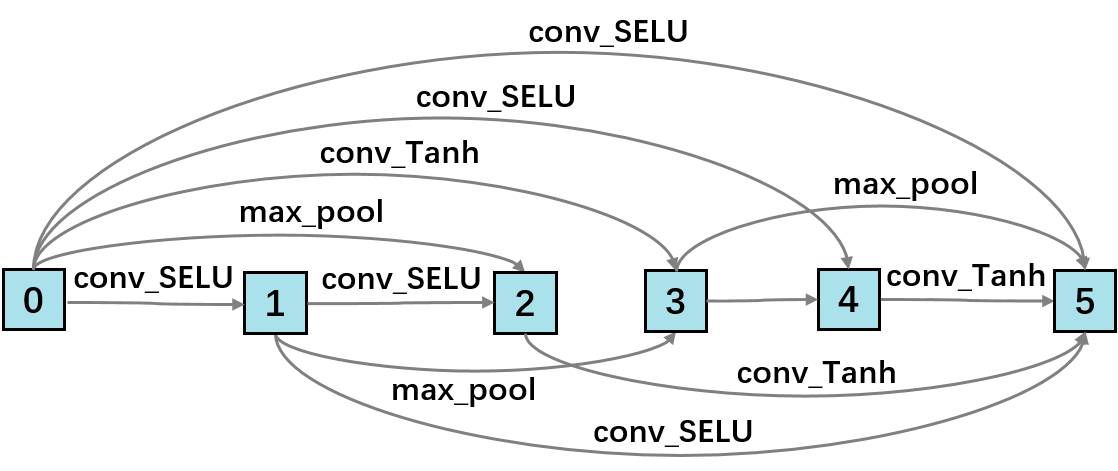}
		\label{cell-cifar}
	}
	\caption{Cell architectures of DPNASNet. }
	\label{cell}
\end{figure*}

\paragraph{Effect of training method.}
To verify the effectiveness of our training method for sampled architectures during the search process, we run our search process by replacing our training method with the ordinary SGD.
Then we use the resulted controller to generate 20 architectures and train all these architectures with DPSGD 10 times on CIFAR-10 for the privacy budget of $(\epsilon, \delta)=(3.0, 1\times10^{-5})$.
The results are presented in Figure \ref{train}. 
Each blue box represents the 10 times results of each architecture and the blue dotted line is the average accuracy of those 20 architectures.
We can see that without using our training method, the resulted architectures still perform better than handcrafted models, e.g. $59.07\%$ test accuracy of CNN-Tanh \cite{ts}, but all the sampled models are less effective than DPNASNet searched by using our training method.

\paragraph{Effect of search strategy.}
To show the effectiveness of the search algorithm, we replace the RL-based search method in our search process with random search and draw the test accuracy of 10 sampled architectures after each epoch in Figure \ref{search}.
We find that, after warm-up (the first 25 epochs), the test accuracy of  models sampled by RL-controller keeps increasingly higher than that from the random sample, which means the search algorithm of our DPNAS is effective.

\paragraph{Transferability of searched architectures.}
The main results for MNIST, FashionMNIST, and CIFAR-10 are obtained by networks searched on these three datasets, respectively.
To show the transferability of searched architectures, 
for each dataset, we construct a model with the cell searched on this dataset and evaluate it on the other two datasets. 
As showen in Table \ref{transfer}, the cell searched on one dataset also perform well when evaluated on other datasets. They can achieve comparable accuracy with the cells searched on the evaluation dataset.
The transferability of NAS searched architectures have been verified in \cite{zoph2018learning}.
Our experiments validate that this transferability still holds when private training is taken into considering in the search process.

\section{Architecture Analysis and Discussions}\label{sec-discuss}
Our searched networks for DPDL show meaningful patterns that are distinct from the networks searched for non-private image classification.
Figure \ref{cell} shows the cell architecture of our searched DPNASNet.
We conduct qualitative and quantitative analysis on the searched architectures by DPNAS and sum up several observations for designing private-learning-friendly networks as following.

\begin{figure}[tbh]
		\centering
		\subfigure[Proportion of activation functions.]{
			\includegraphics[scale=0.35]{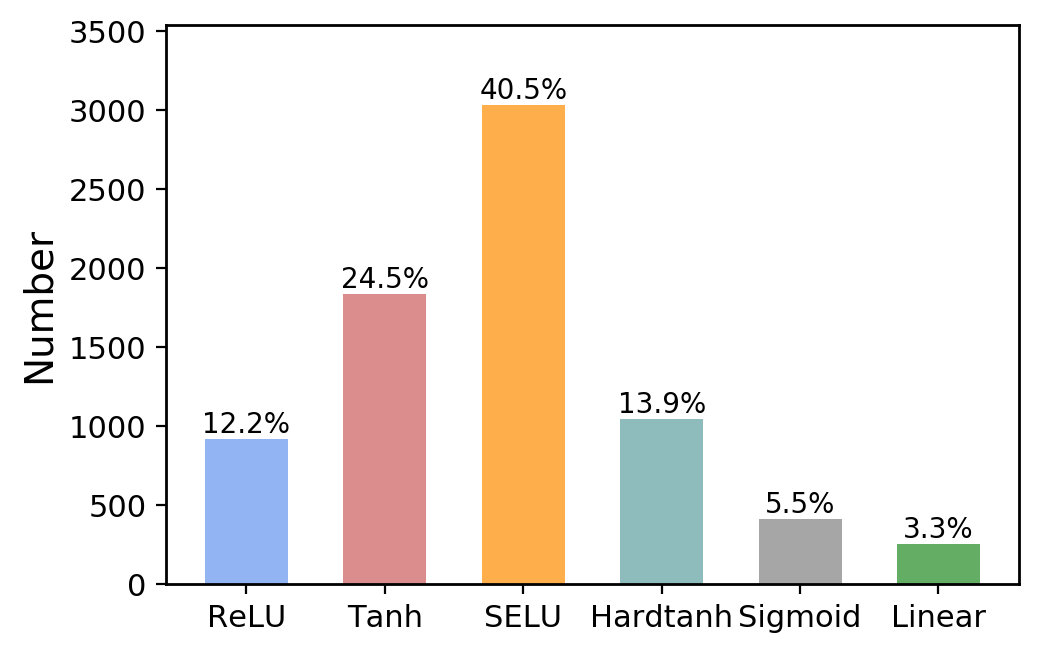}
			\label{act-ratio}
		}
		\subfigure[Proportion of pooling functions.]{
			\includegraphics[scale=0.35]{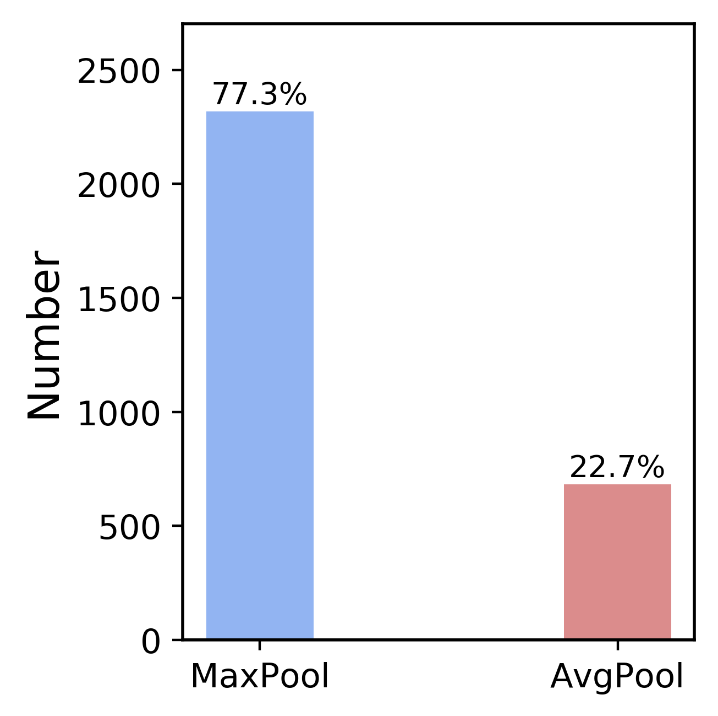}
			\label{pool-ratio}
		}
		
		\subfigure[Comparison between SELU and Tanh.]{
			\includegraphics[scale=0.37]{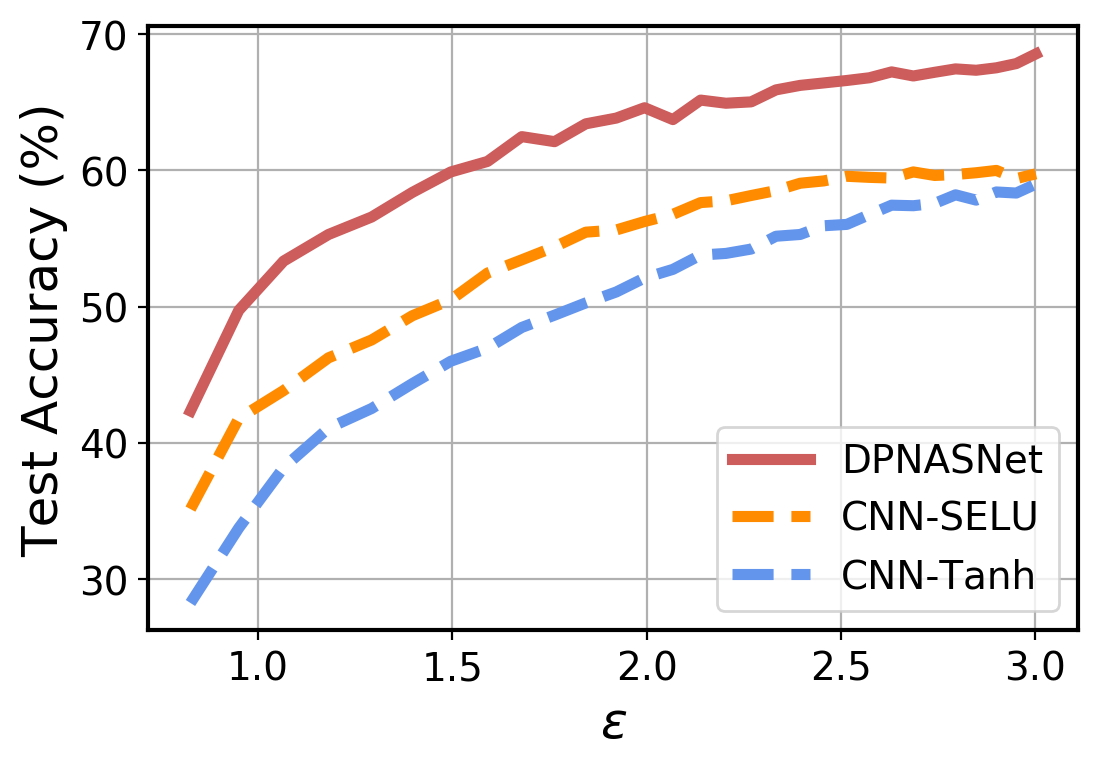}
			\label{act-comp}
		}
		\subfigure[Comparison between AvgPool and MaxPool.]{
			\includegraphics[scale=0.37]{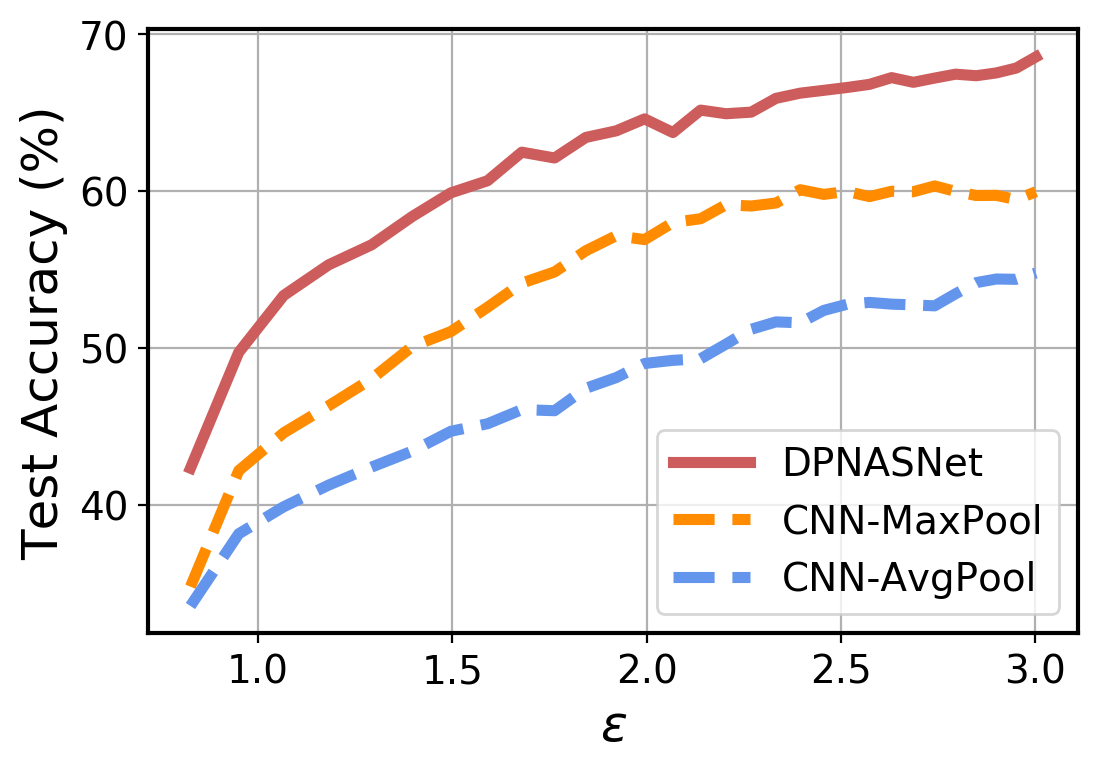}
			\label{pool-comp}
		}
		\caption{Analysis of the generated cells on CIFAR-10. 
			}
		\label{analysis}
\end{figure}

\paragraph{SELU out-performs Tanh for DPDL.} 
From Figure \ref{cell}, we observe that SELU is the most frequently used activation function in the resulted architectures.
To statistically analyze the occurrence frequency of each activation function, we use the trained controller to sample 1000 architectures and count the number of each activation function in these architectures.
As shown in Figure \ref{act-ratio}, the occurrence frequency of SELU is much higher than others.
Tanh also frequently appears and its effectiveness for DPDL has been verified in previous work \cite{ts}.
As SELU appears more frequently than Tanh in our results, we wonder \textit{whether SELU is better than Tanh to construct networks for DPDL}.
To answer this question, we employ the simple CNN model from \cite{ts} with SELU and Tanh as activation function respectively to build two models, CNN-SELU and CNN-Tanh.
From Figure \ref{act-comp}, 
we find that CNN-SELU consistently out-performs CNN-Tanh.
Papernot et al. \cite{ts} argue that the reason for Tanh out-performing ReLU for DPDL is that Tanh is bounded while ReLU is unbounded.
However, we find that SELU is yet efficient for DPDL, although it is also unbounded.
Our intuitive explanation for this is that the activation functions that can retain negative values of their inputs could be more suitable for DPDL.
Here, we try to experimentally verify this intuition.
From Table \ref{act-compare}, we observe that the activation functions that can retain negative values out-perform those only having non-negative values in their outputs.
This result is consistent with our intuition.
We leave the theoretical explanation for future studies.

\begin{table}[thb]  
	\centering  
	\small
	\begin{tabular}{@{}lccc@{}}
		\toprule
		Function  & Has Negative & Bounded  & Test Acc \\ \midrule
		ReLU   & $\times$ & $\times$ & $51.02\%$\\	
		ReLU6  & $\times$ & $\checkmark$ & $53.03\%$ \\
		ELU   & $\checkmark$ & $\times$ & $57.35\%$ \\
		SELU  & $\checkmark$ & $\times$ & $60.16\%$ \\
		Tanh  & $\checkmark$ & $\checkmark$ & $59.21\%$ \\
		HardTanh & $\checkmark$ & $\checkmark$ & $58.46\%$ \\
		LeakyReLU & $\checkmark$ & $\times$ & $59.32\%$ \\
		\bottomrule
	\end{tabular}
	\caption{Comparing the utility of CNN models with different activation layers trained with DPSGD on CIFAR-10.
    }
	\label{act-compare}  
\end{table}

\paragraph{MaxPool performs better than AvgPool.}
From Figure \ref{cell}, we also observe that MaxPool appears more frequently than AvgPool.
We also do a statistic on the occurrence frequency of each pooling function by using a similar statistical method for activation functions.
From Figure \ref{pool-ratio}, we find that the portion of MaxPool used in the resulted architectures is much higher than that of AvgPool. 
Based on this observation, we are curious about \textit{whether MaxPool is better than AvgPool for DPSGD trained models' utility.}
To figure it out, we conduct a comparison on two simple CNN models.
One model employs MaxPool for all its pooling operations, and the other uses AvgPool for all pooling layers.
Both of these two models are trained with DPSGD on CIFAR-10 with the same settings.
Figure \ref{pool-comp} shows that MaxPool is a better selection than AvgPool for the DPSGD trained models.

\section{Conclusion}
We demonstrate that the model architecture has a significant impact on the utility of DPDL.
We then present DPNAS, the first framework of automatically searching models for DPDL.
We specially design a novel search space and propose a DP-aware training method in DPNAS to guide the searched models to be adaptive to DPSGD training. 
The searched model DPNASNet consistently advances the SOTA accuracy on different benchmarks.
Finally, we analyze the generated architectures and provide several new findings of operation selection for designing private-learning-friendly DNNs.

{\small
    \bibliographystyle{plain}
    \bibliography{main}
}


\clearpage
\section*{Appendix}


\subsubsection{Running time of DPNAS search process.}
The running times of search on MNIST with and without DP-aware training are 232 s/epoch and 98 s/epoch, respectively.
In general, the search of DPNAS is slower than the search s without using DP-aware training because the per-example gradient calculation operation of DP-SGD can slow down the training. 
But we note that the running time of searching relies heavily on the implementation of DP-SGD. 
There are some techniques \citep{subramani2020enabling, lee2020scaling} that try to speed up per-example gradient clipping, which are applicable to reduce the search time of DPNAS.

\subsubsection{Inference time of DPNASNet.} We report the inference time of CNN-Tanh, DPNASNet, and DPNASNet-small in Table \ref{inf-time}. We observe that with comparable number of parameters, our architecture DPNASNet-small is a little bit slower than CNN-Tanh but achieves better test accuracy. The reason why our architecture is slower could be that the found cell architectures contain multi-branch, which is not computationally friendly. As we focus on improving the utility of deep learning with DP, the efficiency of resulted models is not the primary goal of this paper. But it could be an interesting direction to consider both DP-learning and efficiency constraints into NAS for searching models that are DP-friendly and efficient. We leave this for future work.
\begin{table}[thbp]  
	\centering  
	\small
	\begin{tabular}{@{}lcccccc@{}}
		\toprule
		Model	&Parameter	&Inference Time & Accuracy \\
        \midrule
        CNN-Tanh	&0.03M	&0.1577s	&$98.10\%$ \\
        DPNASNet-small	&0.03M	&0.2505s	&$98.35\%$\\
        DPNASNet	&0.21M	&0.4703s	&$98.57\%$\\
		\bottomrule
	\end{tabular}
	\caption{Model inference time.
    }
	\label{inf-time}  
\end{table}

\subsubsection{Final architecture varies as a function of the activation or pooling layer.}
In Table \ref{act-func} and Table \ref{pool-func}, we conduct two ablation studies to evaluate the performance of the resulted architecture varies as a function of activation functions and pooling layers, respectively. In the first experiment, we identify the resulted architecture and then replace all the activation functions with the same one type of activation. For example, "only ReLU" indicates we replace all activation functions with ReLU in our searched architecture. We also evaluate another architecture which is obtained by replace each activation function with a random sampled one at each node. The experiment on pooling layers is conducted in a similar way with that on activation function. The results indicate that comparing with searching by taking both architecture topology and component selection into account, decoupling the architecture topology and component selection could lead to sub-optimal results.

\subsubsection{Privacy leakage risk of the search process.}
As the architecture search process directly uses private set, a question we should discuss is whether the architecture search could inadvertently leak private information about the training set.
To answer this question, we first conduct a simple sanity check by evaluating the test accuracy of DPNASNet at initialization. We test DPNASNet for 10 times on CIFAR-10 with different random seed for initialization and the average accuracy of 10 results is $10.62\%$. 
We do not observe the resulted model achieves accuracy that are significantly higher than 10\% (random guess) at initialization.
However, intuitively, the search process on private set still have potential to leak privacy about training data. One way to completely avoid this is to conduct searching on a public dataset and apply the resulted model to learn on private dataset, e.g., search on FashionMNIST and apply on MNIST or CIFAR. As shown in Table 3 of original paper, the cell architectures searched by DPNAS have great transferability, which means the above solution is feasible.

\begin{table}[thbp]  
	\centering  
	\small
	\begin{tabular}{@{}lcccccc@{}}
		\toprule
		Model type	& Accuracy ($\%$) \\
        \midrule
        only ReLU & 66.36 \\
        only SELU & 68.06 \\
        only Tanh & 65.31 \\
        only Linear & 66.81 \\
        only Hardtanh & 64.80 \\
        only Sigmoid & 58.09 \\
        random act & 66.09\\
        \midrule
        DPNASNet & 68.33\\
		\bottomrule
	\end{tabular}
	\caption{Model varies as a function of activation. 
    }
	\label{act-func}  
\end{table}

\begin{table}[thbp]  
	\centering  
	\small
	\begin{tabular}{@{}lccc@{}}
		\toprule
		Model type	& Accuracy ($\%$) \\
        \midrule
        only MaxPool & 68.33 \\
        only AvgPool & 68.47 \\
        random pooling & 66.53 \\
        \midrule
        DPNASNet & 68.33\\
		\bottomrule
	\end{tabular}
	\caption{Model varies as a function of pooling. 
    }
	\label{pool-func}  
\end{table}


\end{document}